\documentclass{egpubl}
\usepackage{eg2019}

\ConferencePaper        %

 \electronicVersion %

\ifpdf \usepackage[pdftex]{graphicx} \pdfcompresslevel=9
\else \usepackage[dvips]{graphicx} \fi

\PrintedOrElectronic

\usepackage{t1enc,dfadobe}
\usepackage{egweblnk}
\usepackage{cite}
\usepackage{amsfonts}
\usepackage{amsmath}
\usepackage{booktabs} %
\usepackage{scalerel}
\usepackage{blindtext}
\usepackage{subcaption}
\usepackage{bm}
\usepackage[dvipsnames,table]{xcolor}
\usepackage[ruled]{algorithm2e} %

\newcommand{\etal}{\textit{et al.}}
\newcommand{\eg}{\textit{e.g.}}
\newcommand{\ie}{\textit{i.e.}}

\definecolor{LightGreen}{HTML}{b9e8cb}
\newcommand{\Dan}[1]{\textcolor{LimeGreen}{[\textbf{Dan}: #1]}} 

\newcommand{\revised}[1]{#1}
 
\newcommand{\removed}[1]{} 
\title[Learning-Based Animation of Clothing for Virtual Try-On]%
{Learning-Based Animation of Clothing for Virtual Try-On}

\author[I. Santesteban, M.A. Otaduy \& D. Casas]
{\parbox{\textwidth}{\centering Igor Santesteban~~~~~~~~~~~~~~   
		Miguel A. Otaduy~~~~~~~~~~~~~~
        Dan Casas
         }
         \\
{\parbox{\textwidth}{\centering Universidad Rey Juan Carlos, Madrid, Spain
        }
 }
}

\begin{document}

	\teaser{
	\vspace{-0.75cm}
        \centering
  		\includegraphics[height=5cm]{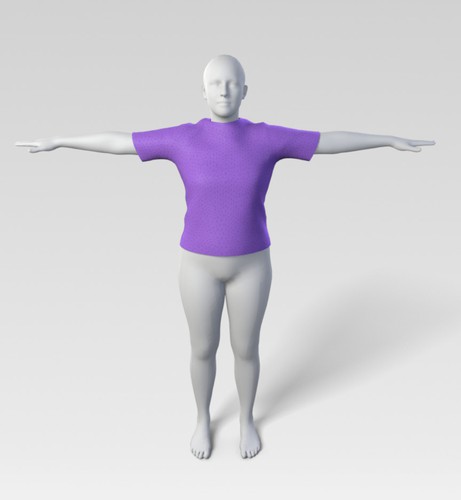}
  		\includegraphics[height=5cm]{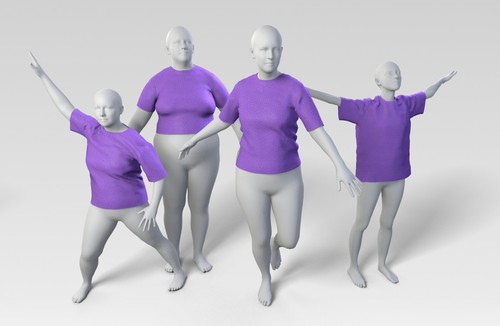}
  		\includegraphics[height=5cm]{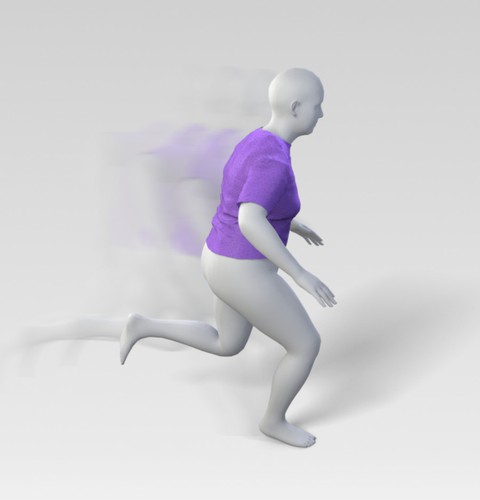}
	   \caption{Given a garment (left), we learn a deformation model that enables virtual try-on by bodies with different shapes and poses (middle). Our model produces cloth animations with realistic dynamic drape and wrinkles at $250$~fps (right).}
	 \label{fig:teaser}
	}
    
	\maketitle
	\begin{abstract}
	This paper presents a learning-based clothing animation method for highly efficient virtual try-on simulation. Given a garment, we preprocess a rich database of physically-based dressed character simulations, for multiple body shapes and animations. Then, using this database, we train a learning-based model of cloth drape and wrinkles, as a function of body shape and dynamics. We propose a model that separates global garment fit, due to body shape, from local garment wrinkles, due to both pose dynamics and body shape. We use a recurrent neural network to regress garment wrinkles, and we achieve highly plausible nonlinear effects, in contrast to the blending artifacts suffered by previous methods. At runtime, dynamic virtual try-on animations are produced in just a few milliseconds for garments with thousands of triangles. We show qualitative and quantitative analysis of results.  

	\end{abstract}  

	\section{Introduction}

Clothing plays a fundamental role in our everyday lives. When we choose clothing to buy or wear, we guide our decisions based on a combination of fit and style. For this reason, the majority of clothing is purchased at brick-and-mortar retail stores, after physical try-on to test the fit and style of several garments on our own bodies. Computer graphics technology promises an opportunity to support online shopping through virtual try-on animation, but to date virtual try-on solutions lack the responsiveness of a physical try-on experience. Beyond online shopping, responsive animation of clothing has an impact on fashion design, video games, and interactive graphics applications as a whole.

\removed{What we wear, and how it fits to our body, does not only influence on how comfortable we feel, but also it is part of the first impression that others see of us. In some cases, clothing even defines our personality.
Consequently, for each piece of cloth we wish to wear, we usually spend a fair amount of time to find the right size, design, and style that best fits our desire.
This trial and error method requires putting on and taking off each piece of cloth, as well as walking around the changing room to test the real feeling of the cloth into our very own body.
The digitalization of this try-on process, enabled by techniques for computer simulation of cloth, can potentially substitute this entire time consuming task. Furthermore, the capability of digitally animating cloth has direct applications in fashion, design, film and video games industries.}

One approach to produce animations of clothing is to simulate the physics of garments in contact with the body.
While this approach has proven capable of generating highly detailed results~\cite{Kaldor:2008:SKC:1360612.1360664,SelleTVCG2009,NarainTOG2012,CLMO14}, it comes at the expense of significant runtime computational cost. On the other hand, it bears no or little preprocessing cost, hence it can be quickly deployed on almost arbitrary combinations of garments and body shapes and motions. To fight the high computational cost, interactive solutions sacrifice accuracy in the form of coarse cloth discretizations, simplified cloth mechanics, or approximate integration methods. Continued progress on the performance of solvers is bringing the approach closer to the performance needs of virtual try-on~\cite{pscc18}. 

An alternative approach for cloth animation is to train a data-driven model that computes cloth deformation as a function of body motion~\cite{Wang2010,deAguiar2010}.
This approach succeeds to produce plausible cloth folds and wrinkles when there is a strong correlation between body pose and cloth deformation.
However, it struggles to represent the nonlinear behavior of cloth deformation and contact in general.
Most data-driven methods rely to a certain extent on linear techniques, hence the resulting wrinkles deform in a seemingly linear manner (\eg, with blending artifacts) and therefore lack realism. 

\removed{%
In the last decade, data-driven approaches have been successful in reducing these compromises. Initial attempts enrich low resolution simulations with fine-scale details from precomputed data~\cite{Wang2010}, or skip the simulation altogether by using a simple regression model and a linear subspace of the clothing~\cite{deAguiar2010}. The key insight about these methods is that they take advantage of the strong correlation between body pose and clothing deformation.
Many more data-driven approaches have been proposed since then \cite{Hahn2014, Guan2012, Kim2013, Xu2014, yangECCV2018}, but most of them still rely to a certain extent on linear techniques. 
Since the behaviour of cloth is highly nonlinear, the use of linear regression models or linear dimensionality reduction methods (\eg ~PCA) usually results in wrinkles that 
deform in a seemingly linear manner and therefore lack realism.}

Most previous data-driven cloth animation methods work for a given garment-avatar pair, and are limited to representing the influence of body pose on cloth deformation.
In virtual try-on, however, a garment may be worn by a diverse set of people, with corresponding avatar models covering a range of body shapes.
In this paper, we propose a learning-based method for cloth animation that meets the needs of virtual try-on, as it models the deformation of a given garment as a function of body motion {\em and} shape.
Other methods that account for changes in body shape do not deform the garment in a realistic way, and either resize the garment while preserving its style~\cite{Guan2012,Brouet:2012:DPG:2185520.2185532}, or retarget cloth wrinkles to bodies of different shapes~\cite{Pons-Moll:Siggraph2017,laehner2018deepwrinkles}.

\removed{Clothing does not only deform due to body pose, but also due to body shape. However, existing data-driven methods either ignore human shape variation \cite{Kim2013, Xu2014, Hahn2014} or, again, represent it with linear models~\cite{Neophytou3DV13,yangECCV2018}.
Alternative methods use wrinkle transfer techniques, sometimes referred as retargeting, ~\cite{Pons-Moll:Siggraph2017,laehner2018deepwrinkles} to copy the deformations of garments across different body shapes, which achieve visually appealing results but far from the real deformations.
Given a garment, our approach enables efficient virtual try-on applications at over 230Hz \Dan{check fps} for arbitrary human shapes, poses and motions. }

We propose a two-level strategy to learn the complex nonlinear deformations of clothing. On one hand, we learn a model of garment fit as a function of body shape. And on the other hand, we learn a model of local garment wrinkles as a function of body shape and motion. Our two-level strategy allows us to disentangle the different sources of cloth deformation.

We compute both the garment fit and the garment wrinkles using nonlinear regression models, \ie, artificial neural networks, and hence we avoid the problems of linear data-driven models.
Furthermore, we propose the use of recurrent neural networks to capture the dynamics of wrinkles. Thanks to this strategy, we avoid adding an external feedback loop to the network, which typically requires a dimensionality reduction step for efficiency reasons~\cite{CO18}.

Our learning-based cloth animation method is formulated as a pose-space deformation, which can be easily integrated into skeletal animation pipelines with little computational overhead.
We demonstrate example animations such as the ones in Figure~\ref{fig:teaser}, with a runtime cost of just $4$ms per frame (more than $1000$x speed-up over a full simulation) for cloth meshes with thousands of triangles, including collision postprocessing. 

To train our learning-based model, we leverage state-of-the-art physics-based cloth simulation techniques~\cite{NarainTOG2012}, together with a parametric human model~\cite{SMPL:2015} and publicly available motion capture data~\cite{CMU,varol17_surreal}. 
In addition to the cloth animation model, we have created a new large dataset of dressed human animations of varying shapes and motions.

	\section{Related Work}
\label{sec:prev}

\paragraph*{Fast Cloth Simulation.}

Physics-based simulation of clothing entails three major processes: computation of internal cloth forces, collision detection, and collision response; and the total simulation cost results from the combined influence of the three processes.
One attempt to limit the cost of simulation has been to approximate dynamics, such as in the case of position-based dynamics~\cite{10.1111:cgf.12346}. While approximate methods produce plausible and expressive results for video game applications, they cannot transmit the realistic cloth behavior needed for virtual try-on.

Another line of work, which tries to retain simulation accuracy, is to handle efficiently both internal forces and collision constraints during time integration.
One example is a fast GPU-based Gauss-Seidel solver of constrained dynamics~\cite{Fratarcangeli:2016:VPG:2980179.2982437}.
Another example is the efficient handling of nonlinearities and dynamically changing constraints as a superset of projective dynamics~\cite{overby2017admmpd}.
Very recently, Tang \etal~\cite{pscc18} have designed a GPU-based solver of cloth dynamics with impact zones, efficiently integrated with GPU-based continuous collision detection.

A different approach to speed up cloth simulation is to apply adaptive remeshing, focusing simulation complexity where needed~\cite{NarainTOG2012}.
Similar in spirit, Eulerian-on-Lagrangian cloth simulation applies remeshing with Eulerian coordinates to efficiently resolve the geometry of sharp sliding contacts~\cite{Weidner:2018:ECS:3197517.3201281}.
\begin{figure*}
	\includegraphics[width=\linewidth]{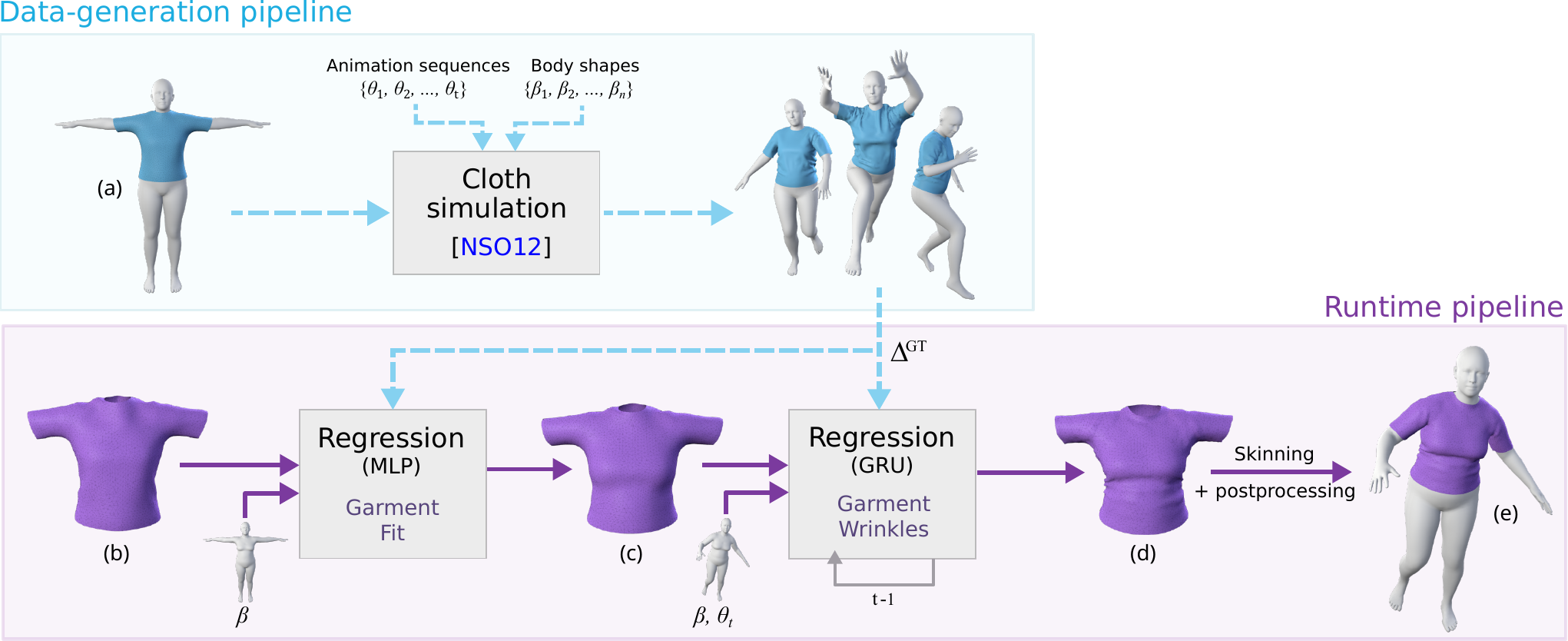}
	\caption{Overview of our preprocessing and runtime pipelines. As a preprocess, we generate physics-based simulations of multiple animated bodies wearing the same garment. At runtime, our data-driven cloth deformation model works by computing two corrective displacements on the unposed garment: global fit displacements dependent on the body's shape, and dynamic wrinkle displacements dependent on the body's shape and pose. Then, the deformed cloth is skinned on the body to produce the final result.
	}
	\label{fig:pipeline}
\end{figure*}
\paragraph*{Data-Driven Models.}
\revised{Multiple works, both well established~\cite{lewis2000pose,sloan2001shape} and recent~\cite{bailey_TOG_2018,CO18}, propose to} model surface deformations as a function of pose. \revised{Similar to them,} some existing data-driven methods for clothing animation also use the underlying kinematic skeletal model to drive the garment deformation \cite{kim2008drivenshape,Wang2010,Guan2012,Xu2014,Hahn2014}.
Kim and Vendrovsky~\cite{kim2008drivenshape} first introduced a pose-space deformation approach that uses a skeletal pose as subspace domain.
Hahn \etal~\cite{Hahn2014} went one step further and performed cloth simulation in pose-dependent dynamic low-dimensional subspaces constructed from precomputed data.
Wang \etal~\cite{Wang2010} used a precomputed database to locally enhance a low-resolution clothing simulation based on joint proximity.

Other methods produce detailed cloth animations by augmenting coarse simulations with example-based wrinkle data.
Rohmer \etal~\cite{rohmer2010animation} used the stretch tensor of a coarse animation output as a guide for wrinkle placement.
Kavan \etal~\shortcite{Kavan2011} used example data to learn an upsampling operator that adds fine details to a coarse cloth mesh.
Zurdo \etal~\cite{Zurdo2013} proposed a mapping between low and high-resolution simulations, employing tracking constraints~\cite{Bergou:2007:TTD} to establish a correspondence between both resolutions.
\revised{Saito \etal~\shortcite{Shunsuke2014} proposed an upsampling technique that adds physically feasible microscopic detail to coarsened meshes by considering the internal 	strain at runtime.}
More recently, Oh \etal \cite{Oh2018} have shown how to train a deep neural network to upsample low-resolution cloth simulations.

A different approach for cloth animation is to approximate full-space simulation models with coarse data-driven models.
James and Fatahalian~\cite{James:2003:PID} used efficient precomputed low-rank approximations of physically-based simulations to achieve interactive deformable scenes.
De Aguiar \etal~\cite{deAguiar2010} learned a low-dimensional linear model to characterize the dynamic behavior of clothing, including an approximation to resolve body-cloth collisions. %
Kim \etal~\cite{Kim2013} performed a near-exhaustive precomputation of the state of a cloth throughout the motion of a character. 
At run-time a secondary motion graph \revised{was} explored to find the closest cloth state for the current pose.
Despite its efficient implementation, the method cannot generalize to new motions.
 Xu \etal~\cite{Xu2014} used a precomputed dataset to mix and match parts of different samples to synthesize a garment mesh that matches the current pose. %
 
As discussed in the introduction, virtual try-on requires cloth models that respond to changes in body pose and shape. However, this is a scarce feature in data-driven cloth animation methods.
 Guan \etal \shortcite{Guan2012} dressed a parametric character and independently modeled cloth deformations due to shape and pose. However, they relied on a linear model that struggles to generate realistic wrinkles, specially under fast motions. Moreover, they accounted for body shape by resizing the cloth model.
 Other works also apply a scaling factor to the garment to fit a given shape, without realistic deformation~\cite{yangECCV2018,Pons-Moll:Siggraph2017,laehner2018deepwrinkles}.

\revised{The variability of the human body has also been addressed in garment design methods. Wang~\shortcite{wang2018rulefree} customized sewing patterns for different body shapes by means of an optimization process. Wang \textit{et al.} \shortcite{garmentdesign_Wang_TOG18} learned a shared shape space that allows the user to directly provide a sketch of the desired look, while the system automatically generates the corresponding sewing patterns and draped garment for different \textit{static} bodies. In contrast, our method aims to estimate the fit of a specific garment (\ie, without altering the underlying sewing patterns) for a wide range of animated bodies.}

\paragraph*{Performance Capture Re-Animation.}
Taking advantage of the recent improvements on performance capture methods \cite{Bradley2008,zhang2017detailed,Pons-Moll:Siggraph2017}, virtual animation of real cloth that has been previously captured (and not simulated) has become an alternative.
Initial attempts fit a parametric human model to the captured 3D scan to enable the re-animation of the captured data, without any explicit cloth layer~\cite{Jain2010MovieReshape,feng2015avatar}.
More elaborate methods extract a cloth layer from the captured 3D scan and fit a parametric model to the actor~\cite{yangECCV2018,Neophytou3DV13,Pons-Moll:Siggraph2017,laehner2018deepwrinkles}. This allows editing the actor's shape and pose parameters while keeping the same captured garment or even changing it. However, re-animated motions lack realism since they cannot predict the nonrigid behavior of clothing under unseen poses or shapes, and are usually limited to copying wrinkles across bodies of different shapes~\cite{Pons-Moll:Siggraph2017,laehner2018deepwrinkles}.

\paragraph*{Image-Based Methods.} Cloth animation and virtual try-on methods have also been explored from an image-based point of view~\cite{scholz2006texture,zhou2012image,hauswiesner2013virtual,hilsmann2013pose,han2018viton}. These methods aim to generate compelling 2D images of dressed characters, without dealing with any 3D model or simulation of any form.
Hilsmann \etal~\cite{hilsmann2013pose} proposed a pose-dependent image-based method that interpolates between images of clothes. 
More recently, Han \etal~\cite{han2018viton} have shown impressive photorealistic results using convolutional neural networks. However, these image-based methods are limited to 2D static images and fixed camera positions, and cannot fully convey the 3D fit and style of a garment.
\revised{
	Other image-based works aim at recovering the full 3D model of both the underlying body shape and garment from monocular video input \cite{yang2018garmentrecovery,rogge2014garment}. This allows garment replacement and transfer, but requires highly complex and expensive pipelines.
	Alternatively, other methods seek to recover the cloth material properties from video~\cite{yang2017learning}, which also opens the door to cloth reanimation.
}

	\section{Clothing Animation}
In this section, we describe our learning-based data-driven method to animate the clothing of a virtual character.
Figure~\ref{fig:pipeline} shows an overview of the method, separating the preprocessing and runtime stages.

In Section~\ref{ssec:clothing} we overview the components of our shape-and-pose-dependent cloth deformation model.
The two key novel ingredients of our model are: (i) a Garment Fit Regressor (Section~\ref{ssec:global_regression}), which allows us to apply global body-shape-dependent deformations to the garment, and (ii) a Garment Wrinkle Regressor (Section \ref{ssec:regression}), which predicts dynamic wrinkle deformations as a function of body shape and pose.

\subsection{Clothing Model}
\label{ssec:clothing}
We denote as $M_{\text{b}}$ a deformed human body mesh, determined by shape parameters $\beta$ (\eg, the principal components of a database of body scans) and pose parameters $\theta$ (\eg, joint angles). We also denote as $M_{\text{c}}$ a deformed garment mesh worn by the human body mesh. A physics-based simulation would produce a cloth mesh $S_{\text{c}}(\beta, \theta)$ as the result of simulating the deformation and contact mechanics of the garment on a body mesh with shape $\beta$ and pose $\theta$. Instead, we approximate $S_{\text{c}}$ using a data-driven model.

Based on the observation that most garments closely follow the deformations of the body, we design our clothing model inspired by the Pose Space Deformation (PSD) literature \cite{lewis2000pose} and subsequent human body models~\cite{anguelov2005scape,feng2015avatar,SMPL:2015}.
We assume that the body mesh is deformed according to a rigged parametric human body model,
\begin{equation}
M_{\text{b}}(\beta, \theta) = W(T_{\text{b}}(\beta, \theta),\beta, \theta,\mathcal{W}_\text{b}), 
\end{equation}
 where $W(\cdot)$ is a skinning function, which deforms an unposed body mesh $T_{\text{b}}(\beta, \theta) \in \mathbb{R}^{3 \times V_{\text{b}}}$ with $V_{\text{b}}$ vertices based on: first, the shape parameters $\beta \in \mathbb{R}^{|\beta|}$, which define joint locations of an underlying skeleton; and second, the pose parameters $\theta\in \mathbb{R}^{|\theta|}$, which are the joint angles to articulate the mesh according to a skinning weight matrix $\mathcal{W}_\text{b}$.
The unposed body mesh may be obtained additionally by deforming a template body mesh $\mathbf{\bar{T}}_{\text{b}}$ to account for body shape and pose-based surface corrections (See, \eg,~\cite{SMPL:2015}).

We propose to model cloth deformations following a similar overall pipeline.
For a given garment, we start from a template cloth mesh $\mathbf{\bar{T}}_{\text{c}} \in \mathbb{R}^{3 \times V_{\text{c}}}$ with $V_{\text{c}}$ vertices, and we deform it in two steps.
First, we compute an unposed cloth mesh $T_{\text{c}}(\beta, \theta)$, and then we deform it using the skinning function $W(\cdot)$ to produce the full cloth deformation.
A key insight in our model is to compute body-shape-dependent garment fit and shape-and-pose-dependent garment wrinkles as corrective displacements to the template cloth mesh, to produce the unposed cloth mesh:
\begin{equation}
T_c(\beta,\theta) = \mathbf{\bar{T}}_{\text{c}} +R_{\scaleto{\text{G}}{4pt}}(\beta) + R_{\scaleto{\text{L}}{4pt}}(\beta,\theta),
\end{equation}
where $R_{\scaleto{\text{G}}{4pt}}()$ and $R_{\scaleto{\text{L}}{4pt}}()$ represent two nonlinear regressors, which take as input body shape parameters and shape and pose parameters, respectively.

The final cloth skinning step can be formally expressed as
\begin{equation}
M_{\text{c}}(\beta, \theta) = W(T_{\text{c}}(\beta, \theta),\beta, \theta,\mathcal{W}_\text{c}).
\end{equation}
\revised{Assuming that most garments closely follow the body,} we define the skinning weight matrix $\mathcal{W}_\text{c}$ by projecting each vertex of the template cloth mesh onto the closest triangle of the template body mesh, and interpolating the body skinning weights $\mathcal{W}_\text{b}$.

The pipeline Figure~\ref{fig:pipeline} shows the template body mesh $\mathbf{\bar{T}}_{\text{b}}$ wearing the template cloth mesh $\mathbf{\bar{T}}_{\text{c}}$ (Figure~\ref{fig:pipeline}-a), and then the template cloth mesh in isolation (Figure~\ref{fig:pipeline}-b), with the addition of garment fit (Figure~\ref{fig:pipeline}-c), with the addition of garment wrinkles (Figure~\ref{fig:pipeline}-d), and the final deformation after the skinning step (Figure~\ref{fig:pipeline}-e).

\begin{figure}[b]
    \centering
    \begin{subfigure}[b]{0.35\linewidth}
  		\includegraphics[width=\linewidth]{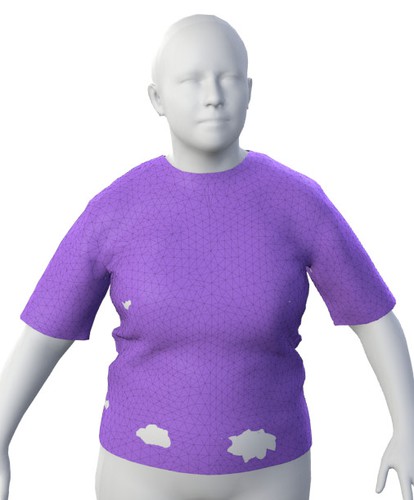}
        \caption{Before}
    \end{subfigure}
    ~~~ 
    \begin{subfigure}[b]{0.35\linewidth}
  		\includegraphics[width=\linewidth]{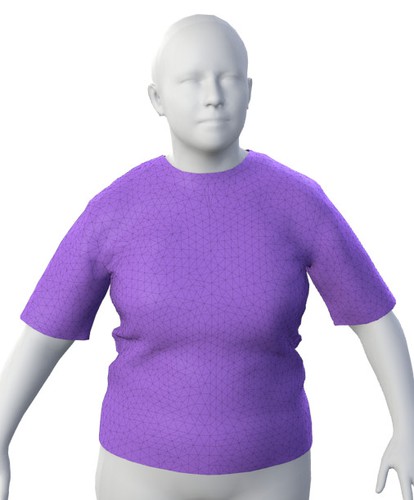}
        \caption{After}
    \end{subfigure}
    ~ 
    \caption{
    	For tight clothing, data-driven cloth deformations may suffer from apparent collisions with the body (left). We apply a simple postprocessing step to push colliding cloth vertices outside the body (right).
    	}
    \label{fig:postprocessing}
\end{figure}

By training regressors with collision-free data, our data-driven model learns naturally to approximate contact interactions, but it does not guarantee collision-free cloth outputs. In particular, when the garments are tight, interpenetrations with the body can become apparent. After the skinning step, we apply a postprocessing step to cloth vertices that collide with the body, by pushing them outside their closest body primitive. An example of collision postprocessing is shown in Figure~\ref{fig:postprocessing}.

\subsection{Garment Fit Regressor}
\label{ssec:global_regression}

Our learning-based cloth deformation model represents corrective displacements on the unposed cloth state, as discussed above. We observe that such displacements are produced by two distinct sources. On one hand, the shape of the body produces an overall deformation in the form of stretch or relaxation, caused by tight or oversized garments, respectively. As we show in this section, we capture this deformation as a static global fit, determined by body shape alone.
On the other hand, body dynamics produce additional global deformation and small-scale wrinkles. We capture this deformation as time-dependent displacements, determined by both body shape and motion, as discussed later in Section \ref{ssec:regression}. We reach higher accuracy by training garment fit and garment wrinkles separately, in particular due to their static vs. dynamic nature.

We characterize static garment fit as a vector of per-vertex displacements $\Delta_{\scaleto{\text{G}}{4pt}} \in \mathbb{R}^{3 \times V_{\text{c}}}$. These displacements represent the deviation between the cloth template mesh $\mathbf{\bar{T}}_{\text{c}}$ and a smoothed version of the simulated cloth worn by the unposed body. Formally, we define the ground-truth garment fit displacements as
\begin{equation}
\Delta^{\scaleto{\text{GT}}{4pt}}_{\scaleto{\text{G}}{4pt}} = \rho(S_{\text{c}}(\beta, \mathbf{0})) - \mathbf{\bar{T}}_{\text{c}},
\label{eq:garment_fit_gt}
\end{equation}
where $S_{\text{c}}(\beta, \mathbf{0}))$ represents a simulation of the garment on a body with shape $\beta$ and pose $\theta = \mathbf{0}$, and $\rho$ represents a smoothing operator.

To compute garment fit displacements in our data-driven model, we use a nonlinear regressor $R_\text{G}: \mathbb{R}^{|\beta|} \rightarrow \mathbb{R}^{3 \times V_{\text{c}}}$, which takes as input the shape of the body $\beta$.
In particular, we implement the regressor $\Delta_{\scaleto{\text{G}}{4pt}} = R_\text{G}(\beta)$ using a single-hidden-layer multilayer perceptron (MLP) neural network.
We train the MLP network by minimizing the mean squared error between predicted displacements $\Delta_{\scaleto{\text{G}}{4pt}}$ and ground-truth displacements $\Delta_{\scaleto{\text{G}}{4pt}}^{\scaleto{\text{GT}}{4pt}}$.

\begin{figure}
	\begin{subfigure}[b]{\columnwidth}
		\includegraphics[width=\columnwidth]{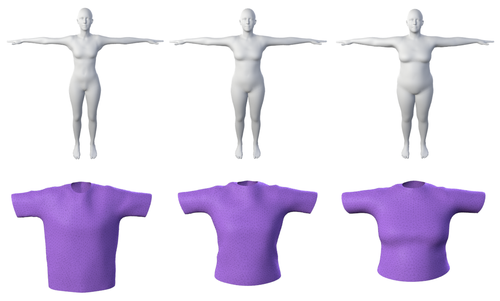}
		\caption{Garment Fit Regression}
		\label{fig:garment_fit_results}
	\end{subfigure}
	
	\vspace{0.2cm}
	\begin{subfigure}[b]{\columnwidth}
		\includegraphics[width=\columnwidth]{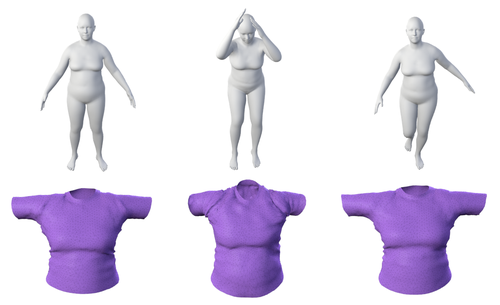}
		\caption{Garment Wrinkle Regression}
		\label{fig:garment_wrinkles_results}
	\end{subfigure}
	\caption{Results of Garment Fit Regression (top) and Garment Wrinkle Regression (bottom), for different bodies and poses. %
	}
\end{figure}

See Figure~\ref{fig:garment_fit_results} for a visualization of the garment  fit regression. Notice how the original template mesh is globally deformed but lacks pose-dependent wrinkles.

\subsection{Garment Wrinkle Regressor}
\label{ssec:regression}

We characterize dynamic cloth deformations (\eg, wrinkles) as a vector of per-vertex displacements $\Delta_{\scaleto{\text{L}}{4pt}} \in \mathbb{R}^{3 \times V_{\text{c}}}$.
These displacements represent the deviation between the simulated cloth worn by the moving body, $S_{\text{c}}(\beta, \theta)$, and the template cloth mesh $\mathbf{\bar{T}}_{\text{c}}$ corrected with the global garment fit $\Delta_{\scaleto{\text{G}}{4pt}}$. We express this deviation in the body's rest pose, by applying the inverse skinning transformation $W^{-1}(\cdot)$ to the simulated cloth. Formally, we define the ground-truth garment wrinkle displacements as
\begin{equation}
\label{eq:garment_wrinkles_gt}
\Delta^{\scaleto{\text{GT}}{4pt}}_{\scaleto{\text{L}}{4pt}} = W^{-1}(S_{\text{c}}(\beta, \theta), \beta, \theta, \mathcal{W}_\text{c})
-
\mathbf{\bar{T}}_{\text{c}} -\Delta_{\scaleto{\text{G}}{4pt}}.
\end{equation}
To compute garment wrinkle displacements in our data-driven model, we use a nonlinear regressor $R_\text{L}: \mathbb{R}^{|\beta| + |\theta|} \rightarrow \mathbb{R}^{3 \times V_{\text{c}}}$, which takes as input the shape $\beta$ and pose $\theta$ of the body.
In contrast to the static garment fit, garment wrinkles exhibit dynamic, history-dependent deformations.
We account for such dynamic effects by introducing recursion within the regressor. In particular, we implement the regressor $\Delta_{\scaleto{\text{L}}{4pt}} = R_\text{L}(\beta, \theta)$ using a Recurrent Neural Network (RNN) based on Gated Recurrent Units (GRU) \cite{cho2014learning}, which has proven successful in modeling dynamic systems such as human pose prediction~\cite{martinez2017human}.
Importantly, GRU networks do not suffer from the well-known vanishing and exploding gradients common in vanilla RNNs~\cite{pascanu2013difficulty}.
Analogous to the MLP network in the garment fit regressor, we train the GRU network by minimizing the mean squared error between predicted displacements $\Delta_{\scaleto{\text{L}}{4pt}}$ and ground-truth displacements $\Delta_{\scaleto{\text{L}}{4pt}}^{\scaleto{\text{GT}}{4pt}}$.%

See Figure \ref{fig:garment_wrinkles_results} for a visualization of the garment wrinkle regression. Notice how the garment obtained in the first step of our pipeline is further deformed and enriched with pose-dependent dynamic wrinkles.

	\section{Training Data and Regressor Settings}
In this section, we give details on the generation of synthetic training sequences and the extraction of ground-truth data to train the regressor networks.
In addition, we discuss the network settings and the hyperparameters used in our results.

\subsection{Dressed Character Animation Dataset}
\label{ssec:dataset}
To produce ground-truth data for the training of the Garment Fit Regressor and the Garment Wrinkle Regressor, we have created a novel dataset of dressed character animations with diverse motions and body shapes.
Our prototype dataset has been created using only one garment, but it can be applied to other garments or their combinations.  

As explained in Section~\ref{ssec:clothing}, our approach relies on the use of a parametric human model.
In our implementation, we have used SMPL~\cite{SMPL:2015}.
We have selected $17$ training body shapes, as follows. For each of the $4$ principal components of the shape parameters $\beta$, we generate $4$ samples, leaving the rest of the parameters in $\beta$ as $0$. To these $16$ body shapes, we add the nominal shape with $\beta = 0$.

As animations, we have selected character motions from the CMU dataset~\cite{CMU}, applied to the SMPL body model~\cite{varol17_surreal}.
Specifically, we have used $56$ sequences containing $7,117$ frames in total (at $30$~fps, downsampled from the original CMU dataset of $120$~fps).
We have simulated each of the $56$ sequences for each of the $17$ body shapes, wearing the same garment mesh (\ie, the T-shirt shown throughout the paper, which consists of $8,710$ triangles).

\begin{figure*}[ht]
	\includegraphics[width=\linewidth]{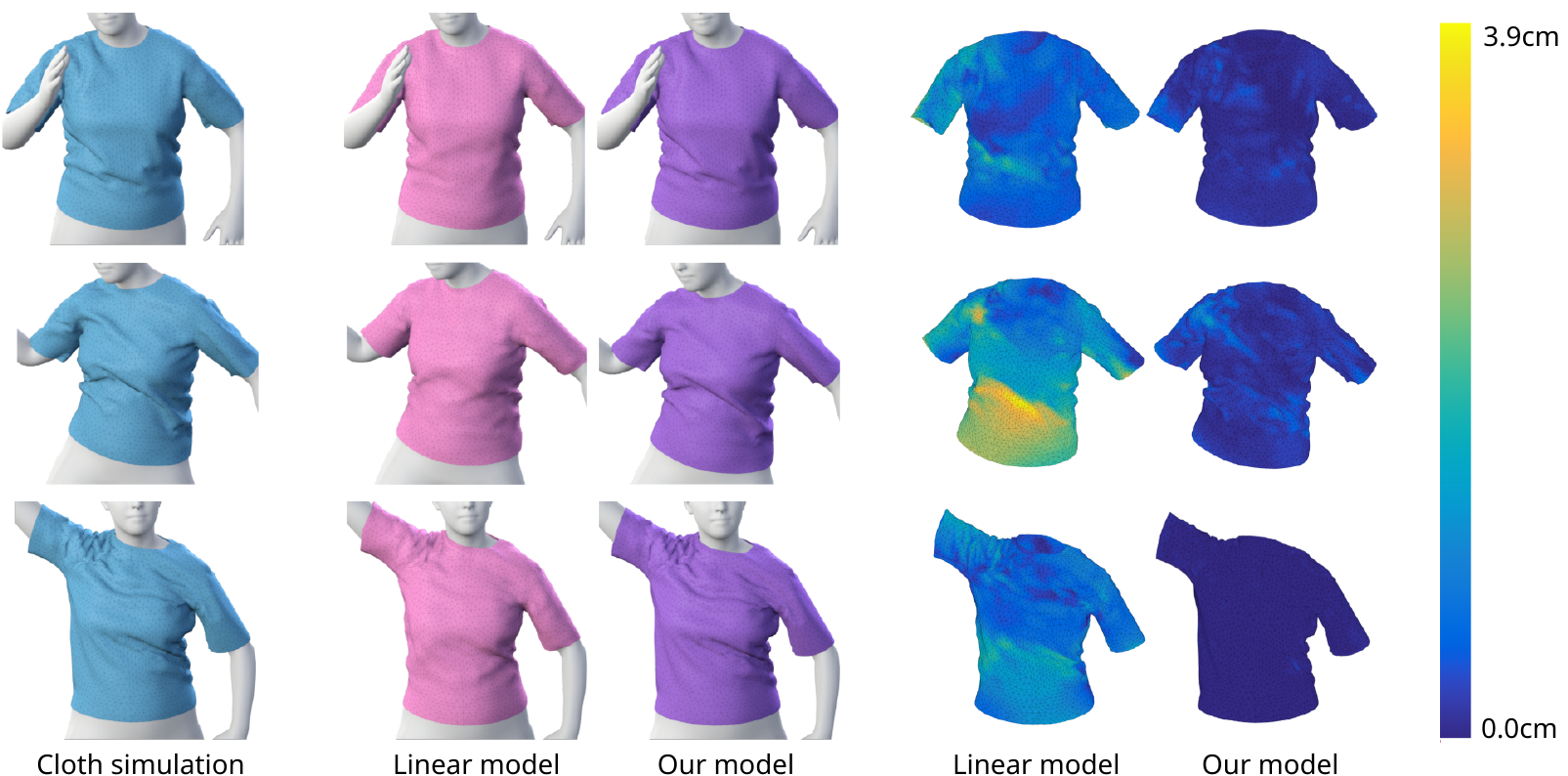}
	\caption{Our nonlinear regression method succeeds to retain the rich and history-dependent wrinkles of the physics-based simulation. Linear regression, on the other hand, suffers blending and smoothing artifacts even on the training sequence shown in the figure.}
	\label{fig:blending}
\end{figure*}

All simulations have been produced using the ARCSim physics-based cloth simulation engine~\cite{NarainTOG2012, Narain:2013:FCA:2461912.2462010}, with remeshing turned off to preserve the topology of the garment mesh.
ARCSim requires setting several material parameters.
In our case, since we are simulating a T-shirt, we have chosen an interlock knit with $60\%$ cotton and $40\%$ polyester, from a set of measured materials~\cite{Wang:2011:DDE}. 
We have executed all simulations using a fixed time step of 3.33ms, with the character animations running at $30$~fps and interpolated to each time step. We have stored in the output database the simulation results from 1 out of every 10 time steps, to match the frame rate of the character animations.
This produces a total of $120,989$ output frames of cloth deformation.

ARCSim requires a valid collision-free initial state. To this end, we manually pre-position the garment mesh \textit{once} on the template body mesh $\mathbf{\bar{T}}_{\text{b}}$. We run the simulation to let the cloth relax, and thus define the initial state for all subsequent simulations. In addition, we apply a smoothing operator $\rho(\cdot)$ to this initial state to obtain the template cloth mesh $\mathbf{\bar{T}}_{\text{c}}$. %

The generation of ground-truth garment fit data requires the simulation of the garment worn by unposed bodies of various shapes. We do this by incrementally interpolating the shape parameters from the template body mesh to the target shape, while simulating the garment from its collision-free initial state. Once the body reaches its target shape, we let the cloth rest, and we compute the ground-truth garment fit displacements $\Delta^{\scaleto{\text{GT}}{4pt}}_{\scaleto{\text{G}}{4pt}}$ according to Equation~\ref{eq:garment_fit_gt}.

Similarly, to simulate the garment on animations with arbitrary pose and shape, we incrementally interpolate both shape and pose parameters from the template body mesh to the shape and initial pose of the animation. Then, we let the cloth rest before starting the actual animation.
The simulations produce cloth meshes $S_{\text{c}}(\beta, \theta)$, and from these we compute the ground-truth garment wrinkle displacements $\Delta^{\scaleto{\text{GT}}{4pt}}_{\scaleto{\text{L}}{4pt}}$ according to Equation~\ref{eq:garment_wrinkles_gt}.

\begin{figure}[t!]
	\includegraphics[width=\linewidth]{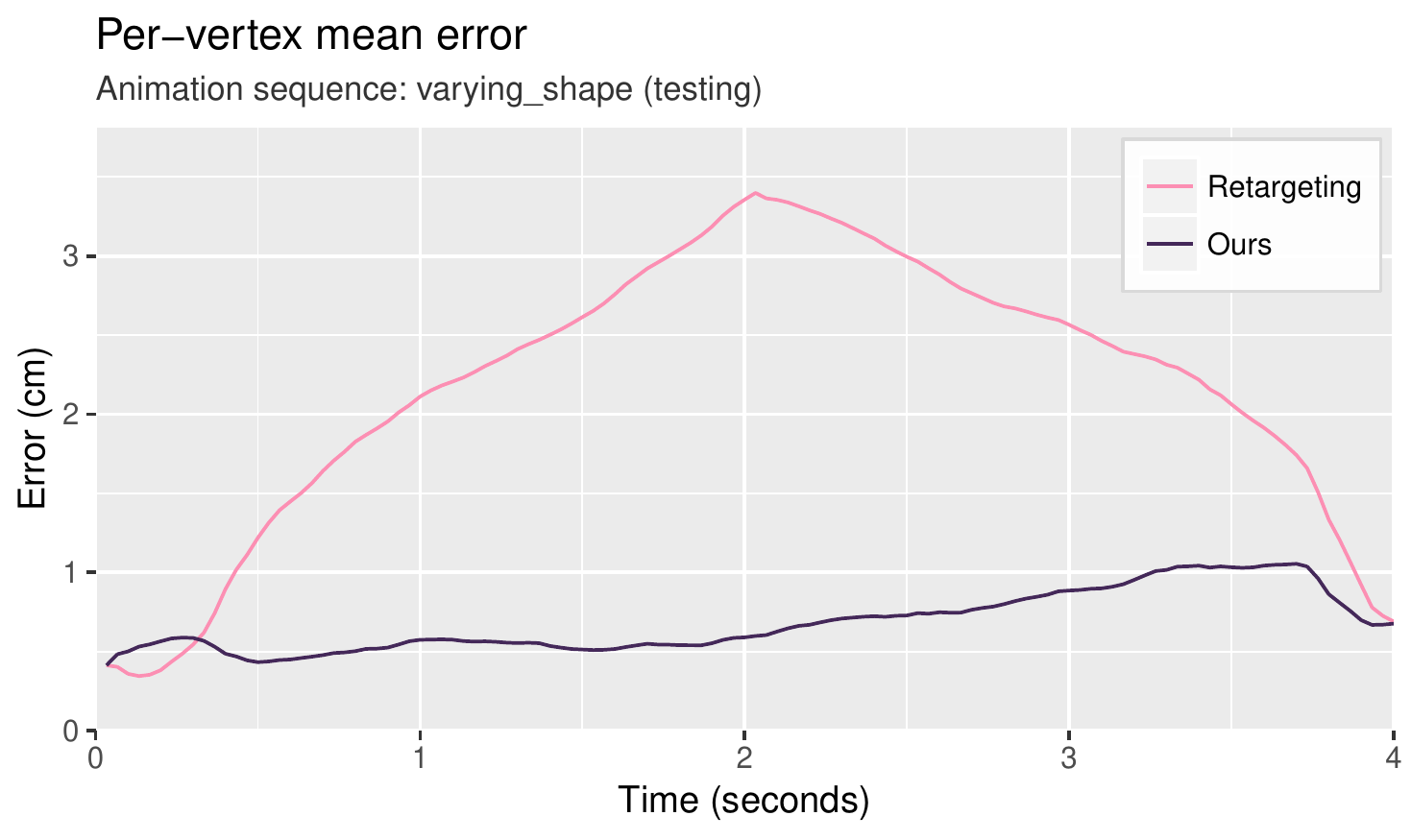}
	\includegraphics[width=\linewidth]{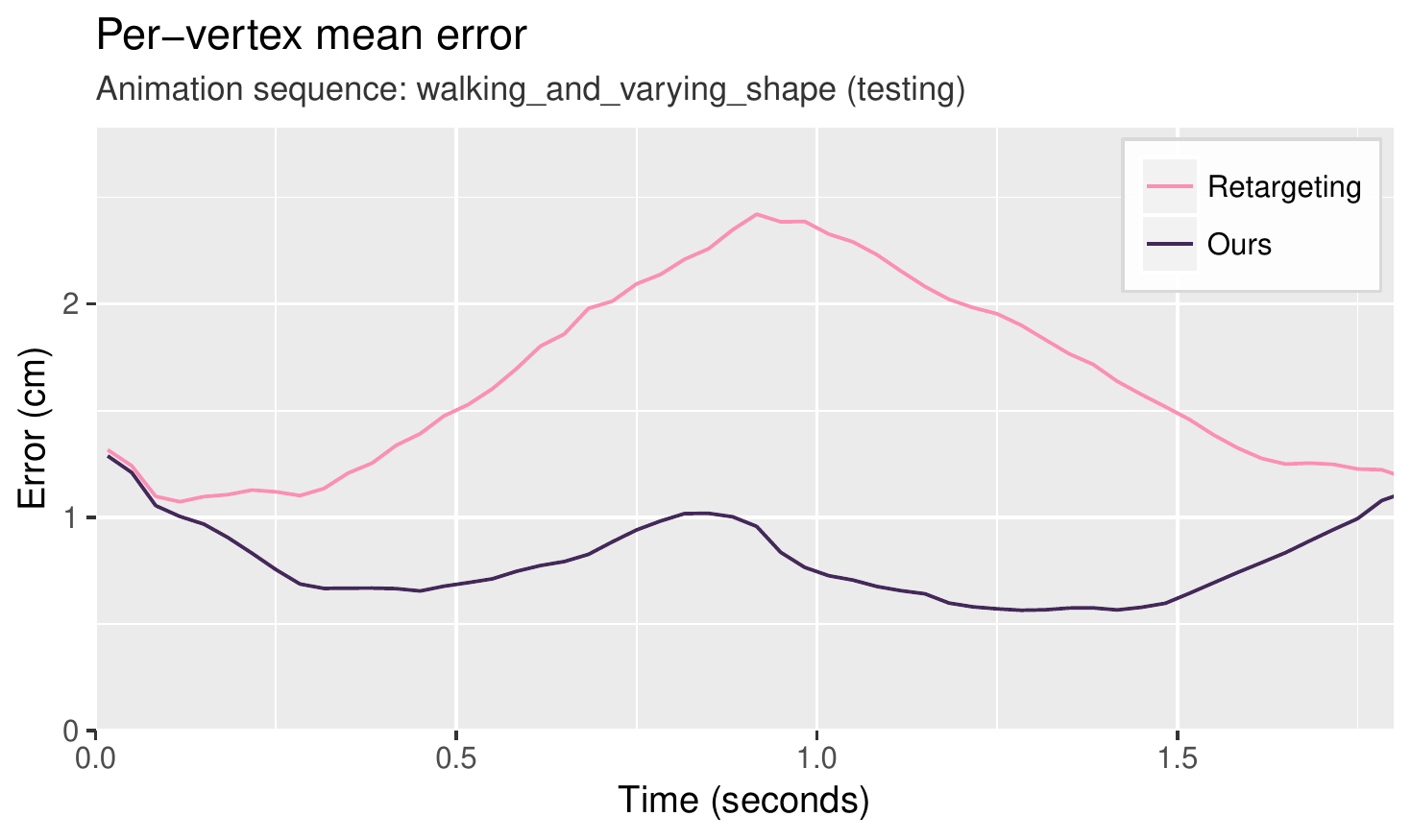}
	\caption{Quantitative evaluation of generalization to new shapes, comparing our method to retargeting techniques~\protect\cite{laehner2018deepwrinkles,Pons-Moll:Siggraph2017}. The top plot shows the error as we increase the body shape to values not used for training, and back, on a static pose (See Figure~\ref{fig:shape_variation}). The bottom plot shows the error as we change both the body shape and pose during a test sequence not used for training.}
	\label{fig:quantiative_ours_vs_retargetting}
\end{figure}

\subsection{Network Implementation and Training}
\label{ssec:training}

We have implemented the neural networks presented in Sections~\ref{ssec:global_regression} and~\ref{ssec:regression} using Tensorflow~\cite{tensorflow2015-whitepaper}. The MLP network for garment fit regression contains a single hidden layer with 20 hidden neurons, which we found enough to predict the global fit of the garment. 
The GRU network for garment wrinkle regression also contains a single hidden layer, but in this case we obtained the best fit of the test data using 1500 hidden neurons. 
In both networks, we have applied dropout regularization to avoid overfitting the training data. Specifically, we randomly disable $20\%$ of the hidden neurons on each optimization step. Moreover, we shuffle the training data at the beginning of each training epoch.  

During training, we use the Adam optimization algorithm \cite{adam2014} for 2000 epochs with an initial learning rate of 0.0001. For the garment fit MLP network, we use for training the ground-truth data from all $17$ body shapes. For the garment wrinkle GRU network, we use for training the ground-truth data from $52$ animation sequences, leaving $4$ sequences for testing purposes.
When training the GRU network, we use a batch size of 128.
Furthermore, to speed-up the training process of the GRU network, we compute the error gradient using Truncated Backpropagation Through Time (TBPTT), with a limit of $90$ time steps.

	\section{Evaluation}
\label{ssec:evaluation}
In this section, we discuss quantitative and qualitative evaluation of the results obtained with our method. 
We compare our results with other state-of-the-art methods, and we demonstrate the benefits of our method for virtual try-on, in terms of both visual fidelity and runtime performance.

\begin{figure}[t!]
	\centering
	\includegraphics[width=\linewidth]{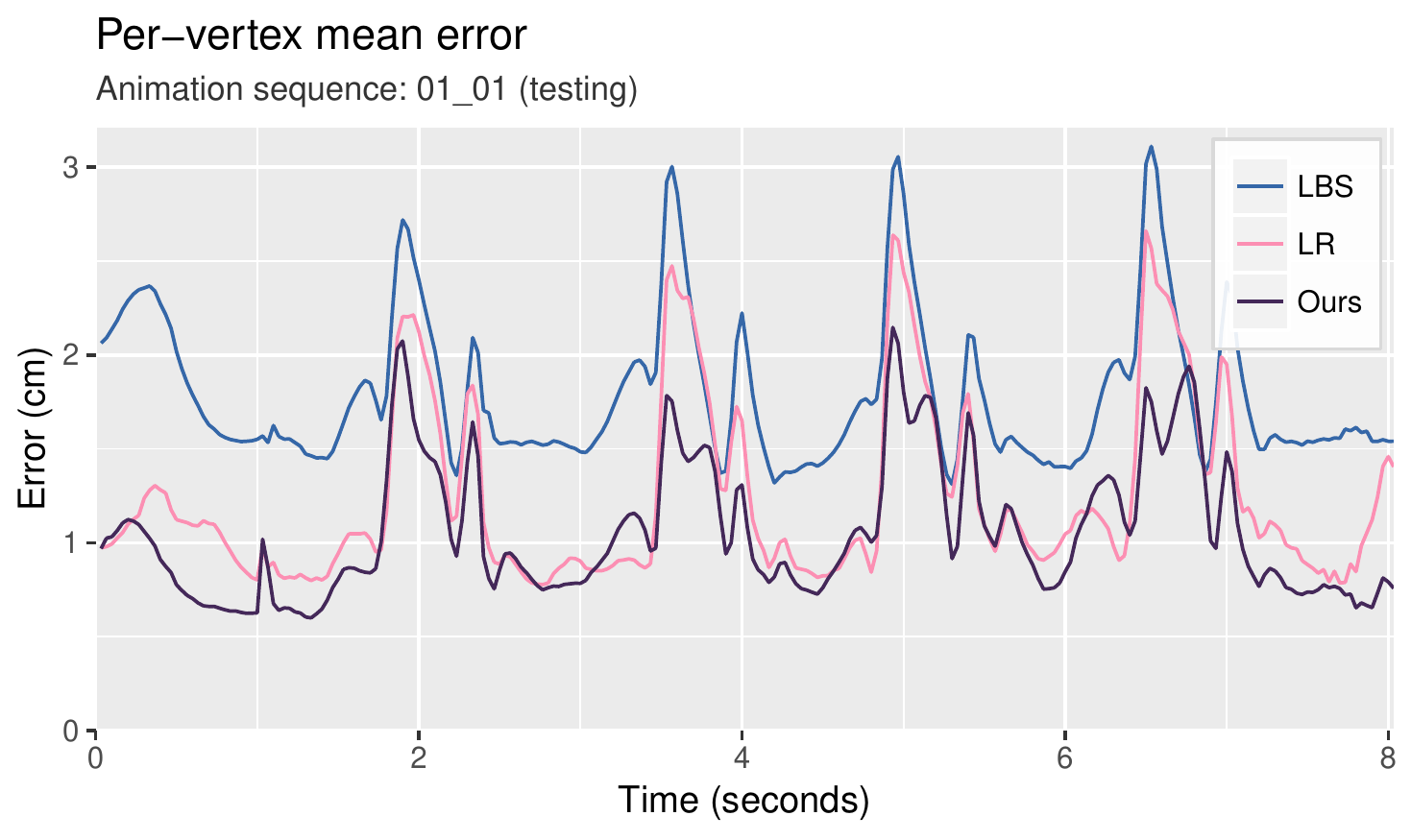}
	\includegraphics[width=\linewidth]{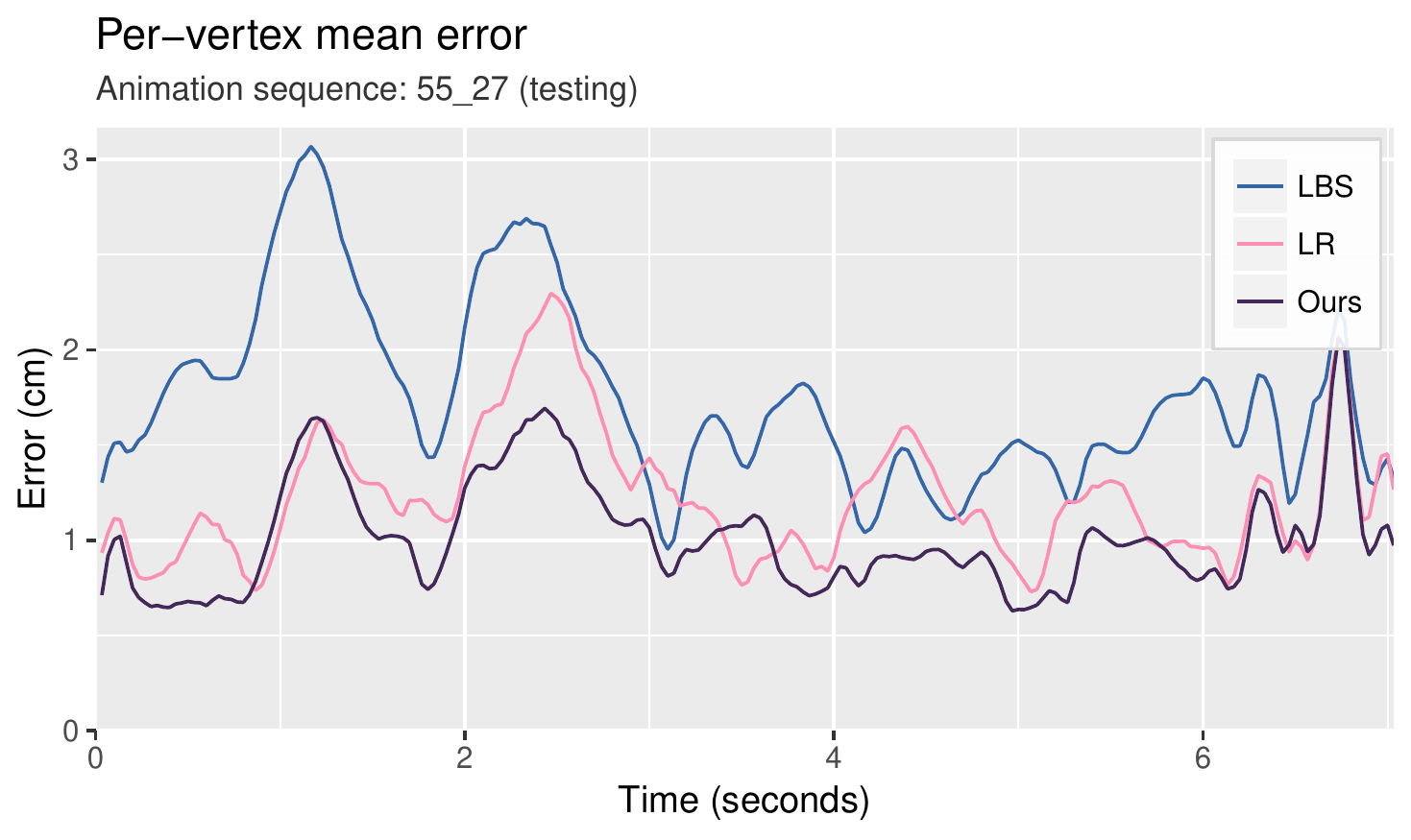}
	\caption{Quantitative evaluation of generalization to new poses, comparing our method to Linear Blend Skinning (LBS) and Linear Regression (LR). 
		Differences with Linear Regression, although visible in the plot, are most evident in the accompanying video.	
	}
	\label{fig:sequence_error}
\end{figure}

\begin{figure*}[t!]
	\centering
	\includegraphics[width=0.95\linewidth]{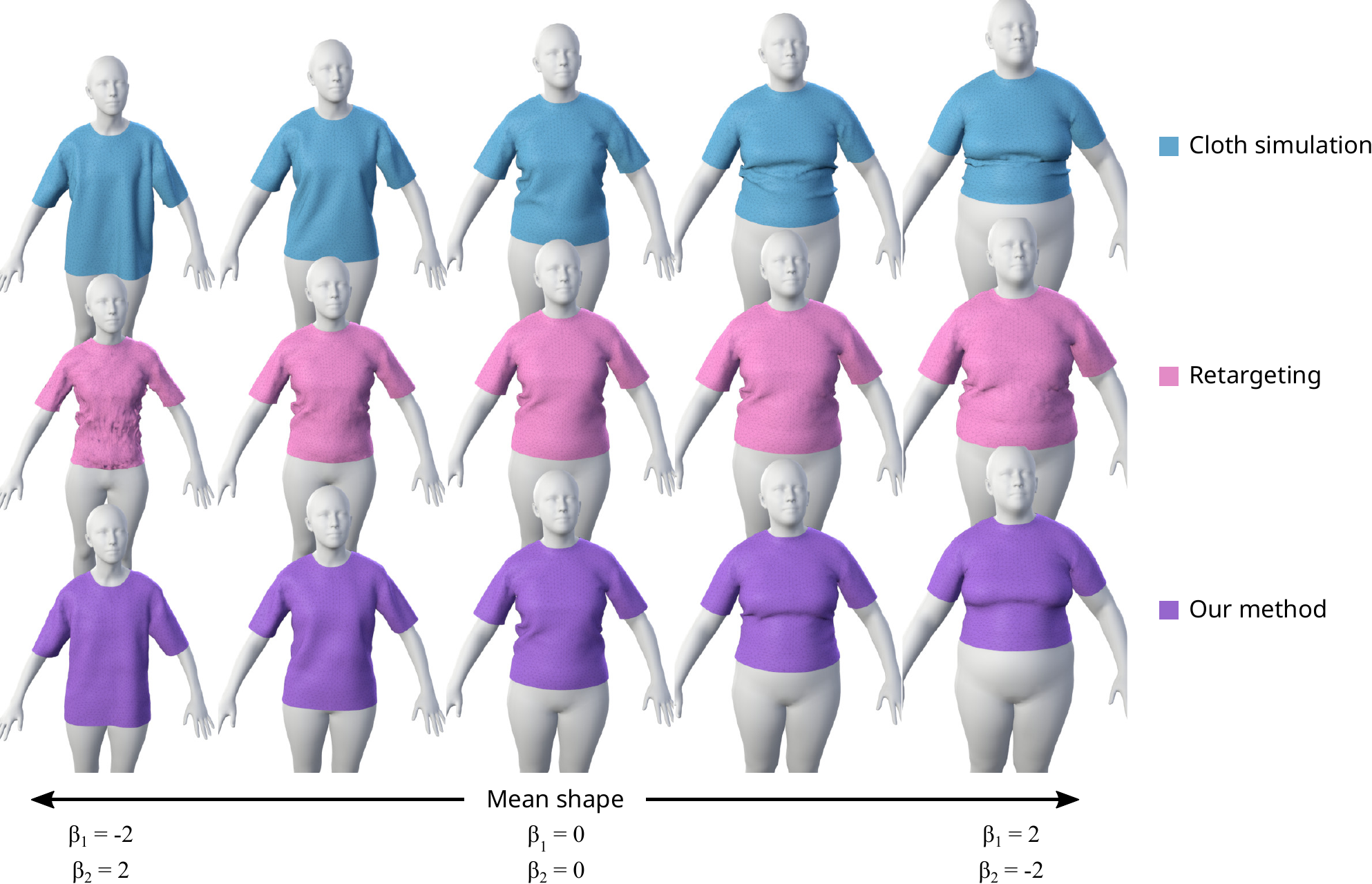}
	\caption{Our method matches qualitatively the deformations of the ground-truth physics-based simulation when changing the body shape beyond training values. In particular, notice how the T-shirt achieves the same overall drape and mid-scale wrinkles. Retargeting techniques~\protect\cite{laehner2018deepwrinkles,Pons-Moll:Siggraph2017}, on the other hand, scale the garment, and suffer noticeable artifacts away from the base shape.}
	\label{fig:shape_variation}
\end{figure*}

\subsection{Runtime Performance}

We have implemented our method on an Intel Core i7-6700 CPU, with a Nvidia Titan X GPU with 32GB of RAM.
Table~\ref{table:runtime} shows average per-frame execution times of our implementation \revised{for different garment resolutions,} including garment fit regression, garment wrinkle regression, and skinning, with and without collision postprocessing. For reference, we also include simulation timings of a CPU-based implementation of full physics-based simulation using ARCSim\cite{NarainTOG2012}.

\begin{table}[h]
    \begin{tabular}{@{\hskip 0mm}lc@{\hskip 3mm}c@{\hskip 5mm}c@{\hskip 0mm}cc@{\hskip 3.5mm}c@{\hskip 3mm}}
    \toprule
    Triangles & 
    \multicolumn{2}{@{\hskip -2mm}c}
    {\begin{tabular}[c]{@{}c@{}}ARCSim\\\cite{NarainTOG2012}\end{tabular}} & 
    \multicolumn{2}{@{\hskip -4mm}c}
    {\begin{tabular}[c]{@{}c@{}}Our method\\(w/o postprocess)\end{tabular}} & 
    \multicolumn{2}{@{\hskip 0mm}c@{\hskip 0mm}}
    {\begin{tabular}[c]{@{}c@{}}Our method\\(w/ postprocess)\end{tabular}} \\
    \cmidrule{2-7} 
    & mean & std & mean & std & mean & std \\
    \midrule
    4,210  & 2909.4  & 1756.8  & 1.47 & 0.31 & 3.39 & 0.30 \\
    8,710  & 5635.4  & 2488.5  & 1.51 & 0.28 & 4.01 & 0.27 \\
    17,710 & 10119.5 & 5849.0  & 2.12 & 0.32 & 5.47 & 0.32 \\
    26,066 & 15964.4 & 4049.3 & 2.40 & 0.33 & 6.87 & 0.30 \\           
    \bottomrule
    \end{tabular}
    \caption{\revised{Per-frame execution times (in milliseconds) of our method for garments of different resolutions, with and without collision postprocessing. Full physics-based simulation times are also provided for reference.}}
    \label{table:runtime}
\end{table}

The low computational cost of our method makes it suitable for interactive applications. Its memory footprint is as follows: 1.1MB for the Garment Fit Regressor MLP, and 108.1MB for the Garment Wrinkle Regressor GRU, both without any compression. 

\subsection{Quantitative Evaluation}
\label{sec:quantitative_evaluation}	

\paragraph*{Linear vs. nonlinear regression.}
In Figure~\ref{fig:blending}, we compare the fitting quality of our nonlinear regression method vs. linear regression (implemented using a single-layer MLP neural network \revised{without nonlinear activation function}), on a training sequence.
While our method retains the rich and history-dependent wrinkles, linear regression suffers smoothing and blending artifacts.

\paragraph*{Generalization to new body shapes.}
In Figure~\ref{fig:quantiative_ours_vs_retargetting}, we quantitatively evaluate the generalization of our method to new shapes (\ie, not in the training set).
We depict the per-vertex mean error on a static pose (top) and a dynamic sequence (bottom), as we change the body shape over time. To provide a quantitative comparison to existing methods, we additionally show the error suffered by   \revised{our implementation of} cloth retargeting~\cite{laehner2018deepwrinkles,Pons-Moll:Siggraph2017}.
As discussed in Section~\ref{sec:prev}, such retargeting methods scale the garment in a way analogous to the body to retain the garment's style. As we show in the accompanying video, even if retargeting produces appealing results, it does not suit the purpose of virtual try-on, and produces larger error w.r.t. a physics-based simulation of the garment.
This is clearly visible in Figure~\ref{fig:quantiative_ours_vs_retargetting}, where the error with retargeting increases as the shape deviates from the nominal shape, while it remains stable with our method.

\paragraph*{Generalization to new body poses.} 
In Figure \ref{fig:sequence_error}, we depict the per-vertex mean error of our method in 2 test motion sequences with constant body shape but varying pose. In particular, we validate our cloth animation results on the CMU sequences 01\_01 and 55\_27~\cite{CMU}, which were excluded from the training set, and exhibit complex motions including jumping, dancing and highly dynamic arm motions.
Additionally, we show the error suffered by two baseline methods for cloth animation. On one hand, Linear Blend Skinning (LBS), which consists of applying the kinematic transformations of the underlying skeleton directly to the garment template mesh. On the other hand, a Linear Regressor (LR) that predicts cloth deformation directly as a function of pose, implemented using a single-layer MLP neural network without nonlinear activation function.
The results demonstrate that our two-step approach, with separate nonlinear regression of garment fit and garment wrinkles, outperforms the linear approach. This is particularly evident in the accompanying video, where the linear regressor exhibits blending artifacts.

\begin{figure}
	\centering
	\includegraphics[width=0.89\columnwidth]{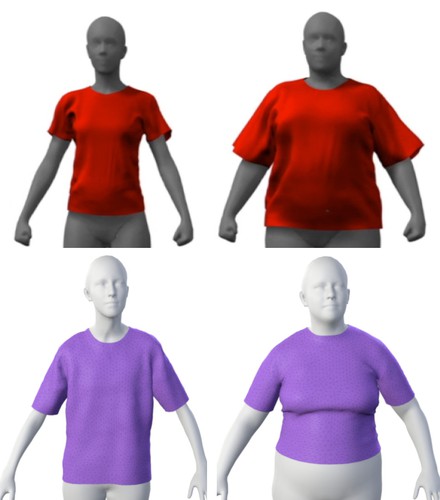}
	\caption{Comparison between DRAPE~\protect\cite{Guan2012} (top) and our method (bottom). DRAPE cannot realistically cope with shape variations, and it is limited to scaling the garment to fit the target shape. In contrast, our method predicts realistically how a garment fits avatars with very diverse body shapes.
	}
	\label{fig:vs_DRAPE}
\end{figure}
\begin{figure}
	\includegraphics[width=\columnwidth]{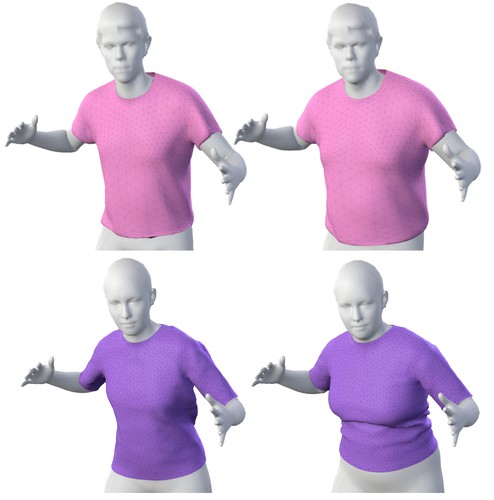}
	\caption{Comparison between ClothCap~\cite{Pons-Moll:Siggraph2017} (top) and our method (bottom). In ClothCap, the original T-shirt (top-left) is obtained using performance capture, and then scaled to fit a bigger avatar. While the result appears plausible for certain applications, it is not suited for virtual try-on. In contrast, our method produces pose- \textit{and} shape-dependent drape and wrinkles, thus enabling a virtual try-on experience. \revised{The skeletal animation used in this comparison and the meshes shown for ClothCap were both provided by the original authors.}}
	\label{fig:clothcap_comparison}
\end{figure}
\begin{figure}
	\includegraphics[width=\linewidth]{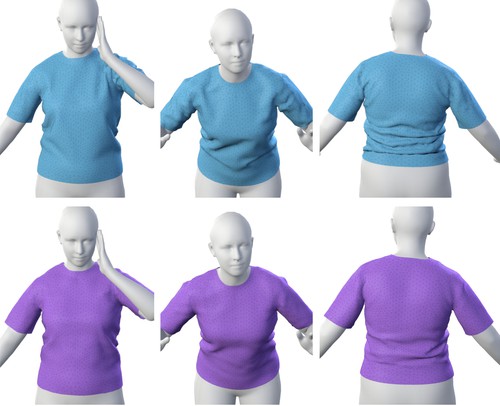}
	\caption{Comparison between a ground-truth physics-based simulation (top) and our data-driven method (bottom), on a test sequence not used for training (01\_01 from \cite{CMU}). 
		Even though our method runs three orders of magnitude faster, it succeeds to predict the overall fit and mid-scale wrinkles of the garment. 
	}
	\label{fig:wrinkles}
\end{figure}
  
\subsection{Qualitative Evaluation}
\label{sec:qualitative_evaluation}

\paragraph*{Generalization to new shapes.}
In Figure~\ref{fig:shape_variation}, we show the clothing deformations produced by our approach on a static pose while changing the body shape over time. We compare results with a physics-based simulation and with \revised{our implementation of} retargeting techniques~\cite{laehner2018deepwrinkles,Pons-Moll:Siggraph2017}.
Notice how our method successfully reproduces ground-truth deformations, including the overall drape (\ie, how the T-shirt slides up the belly due to the stretch caused by the increasingly fat character) and mid-scale wrinkles.

We also compare our method to state-of-the-art data-driven methods that account for changes in both body shape and pose. Figure~\ref{fig:vs_DRAPE} shows the result of DRAPE~\cite{Guan2012} when the same garment is worn by two avatars with significantly different body shapes. DRAPE approximates the deformation of the garment by scaling it such that it fits the target shape, which produces plausible but unrealistic results. In contrast, our method deforms the garment in a realistic manner. 

In Figure~\ref{fig:clothcap_comparison}, we compare our model to ClothCap~\cite{Pons-Moll:Siggraph2017}.
\revised{Admittedly, the virtual try-on scenario considered by Pons-Moll and colleagues differs from ours: while we assume that the virtual garment is provided by a brand and the customer tests the fit of a certain sized garment on a virtual avatar of his/her body, Pons-Moll \textit{et al.} reconstruct the clothing and body shape from 4D scans, and transfer the captured garment to different shapes.}
However, their retargeting lacks realism because cloth deformations are simply copied across different shapes. In contrast, our method produces realistic pose- \textit{and} shape-dependent deformations. 

\begin{figure*}
	\includegraphics[width=\linewidth]{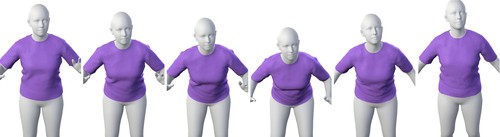}
	\caption{Cloth animation produced by our data-driven method on a test sequence not used for training. Notice how our model successfully deforms the T-shirt and exhibits realistic wrinkles and dynamics.}
	\label{fig:qualitative_results}
\end{figure*}

\paragraph*{Generalization to new poses.}
We visually evaluate the quality of our model in Figure~\ref{fig:wrinkles}, where we compare ground-truth physics-based simulation and our data-driven cloth deformations on a test sequence.
The overall fit and mid-scale wrinkles are successfully predicted using our data-driven model, with a performance gain of three orders of magnitude. 
Similarly, in Figure~\ref{fig:qualitative_results}, we show more frames of a test sequence. Notice the realistic wrinkles in the belly area that appear when the avatar crouches.
Please see the accompanying video for animated results and further comparisons. 

	\section{Conclusions and Future Work}
\label{sec:conclussions}

We have presented a novel data-driven method for animation of clothing that enables efficient virtual try-on applications at over 250 fps. 
Given a garment template worn by a human model, our two-level regression scheme independently models two distinct sources of deformation: garment fit, due to body shape; and garment wrinkles, due to shape and pose.
We have shown that this strategy, in combination with the ability of the regressors to represent nonlinearities and dynamics, allows our method to overcome the limitations of previous data-driven approaches.

We believe our approach makes an important step towards bridging the gap between the accuracy and flexibility of physics-based simulation methods and the computational efficiency of data-driven methods. Nevertheless, there are a number of limitations that remain open for future work.

First, our method requires independent training per \revised{garment size and material}.  %
In particular, \revised{the dataset used in this work consists of a single one-size garment with specific fabric material. Since the material specification can greatly affect the behavior of the cloth, including it as an input to our model would make the wrinkle regression task significantly more challenging. %
Although this particular scenario remains to be tested, using the proposed network architecture would likely result in over-smoothed results.}
Moreover, \revised{multiple independent} garment animations would not capture correctly the interactions between garments. Mix-and-match virtual try-on requires training each possible combination of test garments.

Second, collisions are not fully handled by our method. Our regressors are trained with collision-free data, and therefore our model implicitly learns to approximate contact, but it is not guaranteed to be collision-free. Future work could address this limitation by imposing low-level collision constraints as an explicit objective for the regressor.

Our results show that our method succeeds to predict the overall drape and mid-scale wrinkles of garments, but it smooths excessively high-frequency wrinkles, both spatially and temporally. We wish to investigate alternative methods of recursion to handle accurately both history-dependent draping and highly dynamic wrinkles.

Finally, our model is rooted \revised{in} the assumption that most garments follow closely the body. This assumption may not be valid for loose clothing, and the decomposition of the deformation into a static fit and dynamic wrinkles would not lead to accurate results. It remains to test our method under such conditions.

\revised{\paragraph*{Acknowledgments.} We would like to thank Rosa M. S\'anchez-Banderas and H\'ector Barreiro for their help in editing the supplementary video, and Gerard Pons-Moll and Sergi Pujades for providing us the ClothCap \cite{Pons-Moll:Siggraph2017} meshes. Igor Santesteban was supported by the  Predoctoral Training Programme of the Department of Education of the Basque Government (PRE\_2018\_1\_0307), and Dan Casas was supported by a Marie Curie Individual Fellowship, grant agreement 707326. The work was also funded in part by the European Research Council (ERC Consolidator Grant no. 772738 TouchDesign).}

	\bibliographystyle{eg-alpha-doi}
	\bibliography{santesteban-cloth}
\end{document}